\documentclass[10pt,twocolumn,letterpaper]{article}

\usepackage{cvpr}  

\usepackage{tabularx}
\usepackage{bm}
\usepackage{booktabs}
\usepackage{multirow}
\usepackage{pifont}

\usepackage{float}

\usepackage{soul}
\usepackage{enumitem}
\usepackage{hyperref}

\begin{document}

\title{JDATT: A Joint Distillation Framework for Atmospheric Turbulence Mitigation and Target Detection}

\author{Zhiming Liu, Paul Hill, and Nantheera Anantrasirichai \\
  {Visual Information Laboratory, University of Bristol}, {Bristol}, {UK}}

\maketitle

\begin{abstract}
Atmospheric turbulence (AT) introduces severe degradations—such as rippling, blur, and intensity fluctuations—that hinder both image quality and downstream vision tasks like target detection. While recent deep learning–based approaches have advanced AT mitigation using transformer and Mamba architectures, their high complexity and computational cost make them unsuitable for real-time applications, especially in resource-constrained settings such as remote surveillance. Moreover, the common practice of separating turbulence mitigation and object detection leads to inefficiencies and suboptimal performance. To address these challenges, we propose \textbf{JDATT}—a \textbf{J}oint \textbf{D}istillation framework for \textbf{A}tmospheric \textbf{T}urbulence mitigation and \textbf{T}arget detection. JDATT integrates state-of-the-art AT mitigation and detection modules and introduces a unified knowledge distillation strategy that compresses both components while minimizing performance loss. We employ a hybrid distillation scheme: feature-level distillation via Channel-Wise Distillation (CWD) and Masked Generative Distillation (MGD), and output-level distillation via Kullback–Leibler divergence. Experiments on synthetic and real-world turbulence datasets demonstrate that JDATT achieves superior visual restoration and detection accuracy while significantly reducing model size and inference time, making it well-suited for real-time deployment.
\end{abstract}

\section{Introduction}
\label{sec:intro}

Atmospheric turbulence (AT), caused by temperature differences between the ground and the air, frequently occurs in real-world scenarios. It severely degrades the quality of images and videos by introducing distortions such as rippling, blurring, and intensity fluctuations. These effects pose significant challenges not only to visual fidelity but also to downstream tasks such as object recognition and detection~\cite{Hill2025}. Traditional methods for mitigating atmospheric turbulence distortions—such as deconvolution~\cite{Deledalle:blind:2020} and lucky imaging~\cite{6471221}—struggle with limitations in computational efficiency and tend to introduce artifacts, particularly in dynamic scenes.

In recent years, significant progress has been made in atmospheric turbulence mitigation techniques aimed at enhancing visual quality. Early convolutional neural network (CNN)-based approaches~\cite{gao2019atmospheric} laid the groundwork for more advanced methods such as TMT~\cite{zhang2023imagingatmosphereusingturbulence}, which employs transformer architectures, and MAMAT~\cite{hill2025mamat3dmambabasedatmospheric}, which leverages a 3D Mamba framework~\cite{han2024mamba3denhancinglocalfeatures}. These models have substantially improved image clarity and structural fidelity. However, their high parameter counts and complex architectures result in slow inference speeds, limiting their use in real-time applications. In practice—particularly in time-sensitive domains like remote surveillance—turbulence mitigation alone is insufficient. Accurate and efficient target detection must be jointly addressed.

Although mainstream detection models~\cite{ren2016fasterrcnnrealtimeobject,carion2020endtoendobjectdetectiontransformers} achieve high accuracy, most research prioritizes detection performance over model efficiency and real-time capability. While YOLO~\cite{redmon2016lookonceunifiedrealtime} is known for its real-time detection efficiency, integrating it with  mitigation models in atmospheric turbulence environments introduces substantial computational overhead. This combined framework often compromises real-time performance, particularly in remote monitoring systems with limited computational resources. Consequently, reducing model size and inference time while preserving detection accuracy remains a key challenge.

To address the aforementioned challenges, we propose the first end-to-end joint distillation optimization framework, JDATT (\textbf{J}oint \textbf{D}istillation framework for \textbf{A}tmospheric \textbf{T}urbulence mitigation and \textbf{T}arget detection). JDATT integrates a state-of-the-art atmospheric turbulence mitigation module with a high-performing target detection model under atmospheric turbulence conditions,  optimized jointly via a unified knowledge distillation (KD) strategy to enable model compression with minimal performance degradation.

Specifically, the proposed joint distillation framework adopts a hybrid strategy that combines feature-level and output-level distillation. For feature-level supervision, we employ Channel-Wise Distillation (CWD)~\cite{shu2021channelwiseknowledgedistillationdense} and Masked Generative Distillation (MGD)\cite{yang2022maskedgenerativedistillation}. CWD aligns intermediate channel responses between teacher and student, while MGD introduces region-specific masks to guide the student toward salient areas. At the output level, we apply Kullback–Leibler (KL) divergence~\cite{hinton2015distillingknowledgeneuralnetwork} to match the student’s class probability distributions to those of the teacher, facilitating the transfer of inter-class relationships. These components are complementary: feature-level distillation conveys representational capacity, while output-level distillation refines prediction alignment. Together, they enable efficient knowledge transfer within a compact model architecture.

Experiments on both synthetic and real atmospheric turbulence datasets show that the joint distillation training strategy yields better visual quality in restored videos and higher target detection accuracy compared to separate distillation processes. Our knowledge distillation strategy reduces model size and inference time, demonstrating its effectiveness in turbulence-affected target detection scenarios.
\section{Related Work}

\paragraph{Atmospheric turbulence mitigation.}
Over the past decade, deep learning has driven the development of numerous turbulence mitigation methods. Early approaches were primarily based on CNN architectures~\cite{gao2019atmospheric, mao2021acceleratingatmosphericturbulencesimulation, anan2023atmospheric}, but recent work has demonstrated the superior capability of transformers in capturing complex distortion patterns. TurbNet~\cite{mao2022single} adopts a UNet-like architecture, replacing convolutional layers with transformer blocks. TMT~\cite{zhang2023imagingatmosphereusingturbulence} introduces a two-stage pipeline—de-tilting to correct local shifts and de-blurring to enhance clarity—guided by a multi-scale loss to improve performance under varying turbulence levels. Wo et al.~\cite{Wu:24} enhance local-global context interaction using sliding window-based self-attention and channel attention, inspired by phase distortion and point spread function representations. Physics-inspired models have also gained traction for turbulence removal~\cite{Jaiswal:Physics:2023, Jiang:NeRT:2023}. Diffusion models have shown superior performance on single-image restoration~\cite{Nair:AT-DDPM:2023}, while transformer-based methods remain state-of-the-art for video restoration~\cite{Anantrasirichai:Artificial:2021}. The DATUM framework~\cite{zhang2024spatiotemporalturbulencemitigationtranslational} replicates classical mitigation pipelines~\cite{6471221}, combining deformable attention for frame alignment and residual dense blocks for fusion. Most recently, MAMAT~\cite{hill2025mamat3dmambabasedatmospheric} integrates 3D Mamba within a UNet-like architecture, outperforming both TMT and DATUM on synthetic benchmarks.

\paragraph{Object detection.} 
Object detectors are broadly classified into two-stage and one-stage models. Two-stage detectors, like the R-CNN family~\cite{girshick2014rich}, first generate region proposals and then classify them.  Faster R-CNN~\cite{ren2016fasterrcnnrealtimeobject} improve efficiency by unifying proposal and classification steps. One-stage detectors, such as YOLO~\cite{redmon2016lookonceunifiedrealtime}, predict bounding boxes and class probabilities in a single pass, offering faster performance for real-time tasks like surveillance and autonomous driving. Successive YOLO versions (e.g. \cite{wang2022yolov7,khanam2024yolov11overviewkeyarchitectural}) introduce improvements such as anchor boxes, multi-scale detection, and advanced augmentation strategies. Transformer-based detectors like DETR~\cite{carion2020endtoendobjectdetectiontransformers} replace traditional components with a transformer encoder-decoder and bipartite matching. Extensions like Deformable DETR~\cite{zhu2021deformable} enhance performance on small objects and improve training convergence. To date, only two methods have specifically proposed detector architectures designed to better handle turbulence-induced distortions~\cite{Lau:ATFaceGAN:2020, Hu:Object:2023}.

\paragraph{Knowledge distillation (KD).} Knowledge distillation (KD) is a widely adopted model compression technique that enables a lightweight student network to inherit the performance characteristics of a larger, high-capacity teacher model. Initially introduced by Hinton et al.~\cite{hinton2015distillingknowledgeneuralnetwork}, KD leverages the teacher’s softened output logits to guide the student model with richer supervisory signals than one-hot labels alone. This approach has become especially influential in computer vision tasks where model efficiency is crucial, such as object detection, semantic segmentation, and real-time video analysis.
Existing KD techniques can be broadly categorized into three groups:
{i) Response-based distillation:} The most classic form of KD, in which the student learns to replicate the teacher’s output probability distribution~\cite{hinton2015distillingknowledgeneuralnetwork}.
{ii) Feature-based distillation:} These methods supervise the student using the teacher’s intermediate feature representations or activation maps, often incorporating attention or normalization mechanisms to better align spatial or channel-wise features~\cite{romero2015fitnetshintsdeepnets, zhou2020channeldistillationchannelwiseattention}.
{iii) Relation-based distillation:} Instead of matching raw features, these methods distill the relational knowledge encoded in the teacher’s internal representations, such as pairwise distances, similarities, or graph-based structures among instances or feature channels~\cite{park2019relationalknowledgedistillation}.
 
\section{Methodology}

\begin{figure*}
    \begin{center}
    \includegraphics[width=\linewidth]{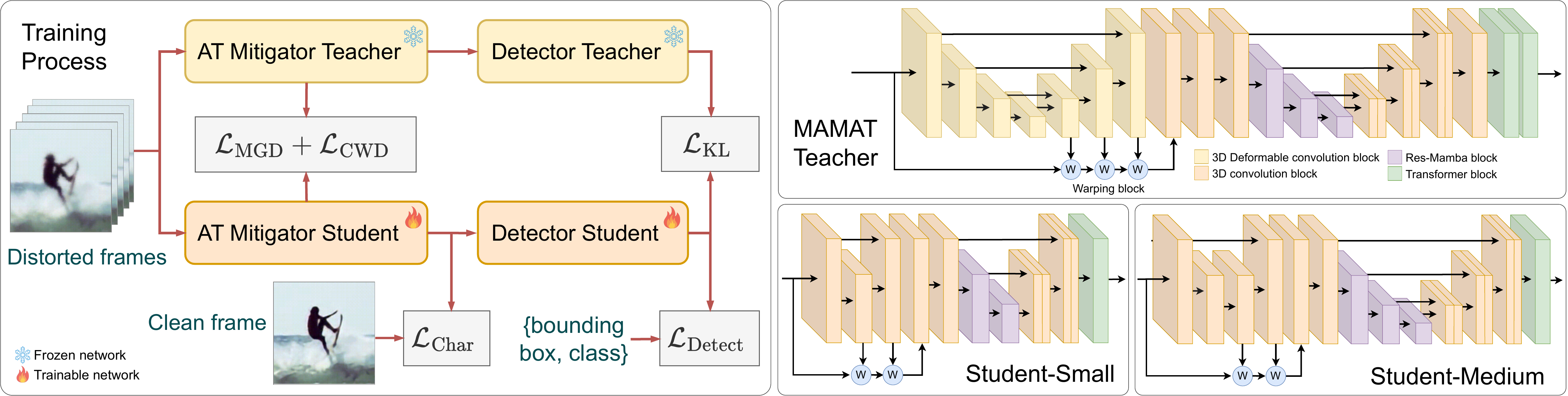}
    \end{center}
    \caption{Overview of the proposed JDATT framework. (Left) Joint knowledge distillation (KD) training pipeline integrating turbulence mitigation and object detection. (Right) Example architectures of the teacher AT mitigator (MAMAT~\cite{hill2025mamat3dmambabasedatmospheric}) and the newly designed lightweight student models in small and medium configurations.}
    \label{fig:distillation}
\end{figure*}

Our proposed JDATT framework is illustrated in Fig.~\ref{fig:distillation} (left), where two pretrained models—serving as teachers for atmospheric turbulence mitigation and target detection—are frozen, while two smaller student models are jointly trained. The objective is joint optimization, enabling mutual reinforcement and performance improvement by simultaneously training multiple related modules within a unified framework. Unlike stage-wise training, which integrates modules after independent training~\cite{yuan2019stagewisetrainingacceleratesconvergence}, our joint training strategy employs a unified loss function to collaboratively update the parameters of all components.

Through end-to-end optimization, the two modules no longer distill knowledge in isolation but instead form a bidirectional feedback loop via joint backpropagation: the turbulence mitigation module learns to generate feature representations that are more conducive to object detection, while gradient signals from the detection module guide the restoration module to preserve visual features critical for recognition. This joint training strategy not only retains the lightweight design of the model but also mitigates accuracy degradation associated with separate distillation.

As illustrated in Fig.~\ref{fig:distillation} (left), the proposed joint distillation framework is optimized using multiple loss functions. In the student pathway, the restored output from the AT mitigation module is supervised by a reconstruction loss to approximate the clean image, while standard detection losses are applied to ensure accurate bounding box regression and object classification. Knowledge distillation is enforced through feature-level supervision using Channel-Wise Distillation (CWD) and Masked Generative Distillation (MGD), guiding the student to capture salient representations learned by the teacher. The restored output is then forwarded to the detection module, where Kullback–Leibler (KL) divergence is employed at the output level to align the student’s prediction distribution with that of the teacher.

Fig.~\ref{fig:distillation} (right) illustrates the architecture of our student models for the AT mitigation module. With supervision from the teacher model during training, the student architecture is designed to be more lightweight. Specifically, the student networks employ a shallower structure, with several convolutional layers removed to reduce complexity. Furthermore, although deformable convolutions have proven highly effective in recent AT mitigation models~\cite{zhang2024spatiotemporalturbulencemitigationtranslational, hill2025mamat3dmambabasedatmospheric}, they are replaced with standard convolutional layers in the student models to improve inference efficiency while maintaining competitive performance through distillation.

\subsection{Reconstruction loss}

Atmospheric turbulence introduces spatially varying distortions and outliers, which pose challenges to conventional pixel-wise loss functions. To address this, we adopt the Charbonnier loss, a robust loss formulation that effectively combines the advantages of both $\ell_1$ and $\ell_2$ norms. Unlike $\ell_2$ loss, which is sensitive to outliers, or $\ell_1$ loss, which may lack smoothness near zero, the Charbonnier loss offers a smooth approximation that retains robustness to outliers while maintaining differentiability:
\begin{equation}
L_\text{Char}(x, y) = \sqrt{(x - y)^2 + \epsilon^2},
\label{Loss}
\end{equation}
where $x$ and $y$ denote the predicted and ground truth values, respectively, and $\epsilon$ is a small constant (typically $1 \times 10^{-3}$) introduced to ensure numerical stability. The Charbonnier loss enables stable training under noisy conditions, making it especially suitable for turbulence mitigation tasks where subtle deviations and noise are prevalent.

\subsection{Detection loss}

The detection loss combines three distinct loss components: box regression loss ($\mathcal{L}_\text{boxes}$), generalized IoU loss ($\mathcal{L}_\text{GIoU}$)~\cite{rezatofighi2019generalized}, and classification loss ($\mathcal{L}_\text{labels}$). Specifically, $\mathcal{L}_\text{boxes}$ is an $\ell_1$ regression loss for predicting box locations and sizes. $\mathcal{L}_\text{GIoU}$ penalizes incorrect box overlap by employing generalized Intersection-over-Union (IoU). The classification component, $\mathcal{L}_\text{labels}$, employs a Binary Cross-Entropy (BCE) loss computed from the predicted class probabilities $p_\text{ce}$, modulated by IoU scores~\cite{Cai_2024_BMVC}. It is defined as:
\begin{equation} 
\begin{split}
\mathcal{L}_\text{labels} &= - \frac{1}{N_{pos}+N_{neg}} [ \sum^{N_{pos}} t\log p_\text{ce} \\ 
& + (1-t) \log (1 - p_\text{ce}) + \sum^{N_{neg}}  p_\text{ce}^\gamma \log (1 - p_\text{ce}) ],
\end{split}
\label{eq:Loss_label}
\end{equation} 
where $N_{pos}$ and $N_{neg}$ denote positive and negative box counts respectively, $\gamma$ (set to 2) balances the weighting, and $t=p_\text{ce}^\alpha \text{IoU}^{(1-\alpha)}$ smoothly transitions targets with an exponential decay controlled by $\alpha = 0.25$.

\subsection{Distillation losses}

\subsubsection{Channel-Wise Distillation Loss (CWD).}
CWD loss~\cite{shu2021channelwiseknowledgedistillationdense} is an effective knowledge distillation technique designed specifically for lightweight models. Unlike conventional distillation methods that directly replicate weights, CWD loss encourages the student model to learn meaningful intermediate feature representations from the teacher model. Specifically, CWD aligns feature distributions between student and teacher models at the channel level, enhancing the student's ability to capture critical information without compromising its lightweight architecture. Consequently, the student model achieves improved recognition performance by selectively absorbing valuable features from the teacher model.
The MGD Loss is formulated as follows~\cite{shu2021channelwiseknowledgedistillationdense}:
\begin{equation}
    \mathcal{L}_\text{MGD}(s, t) = EDT(r) \, CD(s, t) + GKD(s, t) + CE(s, y)
\end{equation}
where \(s\) and \(t\) denote the student and teacher predictions, respectively, \(CD(s, t)\) represents the Channel Distillation loss, \(GKD(s, t)\) denotes the Global Knowledge Distillation loss, \(CE(s, y)\) is the standard Cross-Entropy loss between the student prediction and the ground truth label \(y\), and \(EDT(r)\) is a balancing coefficient depending on the hyperparameter \(r\).

\subsubsection{Mask Generation Distillation Loss (MGD).}
MGD loss~\cite{yang2022maskedgenerativedistillation} guides student models to effectively learn features from teacher models via partial masking. Unlike conventional distillation methods that directly compare entire feature maps, MGD employs a random mask to selectively align features only in unmasked regions, enabling the student model to focus on salient areas. Specifically, during training, MGD utilizes a generative module (e.g., a small convolutional network) to reconstruct masked student features, which are then compared to corresponding teacher features to compute the loss. Consequently, the student not only captures critical features from the teacher but also improves its robustness across diverse spatial contexts, making MGD particularly beneficial for model compression tasks. The formula for MGD loss is shown below~\cite{yang2022maskedgenerativedistillation}:
\begin{equation} 
\begin{split}
\mathcal{L}_\text{MGD}(S, T) &= \sum_{l=1}^{L} \sum_{k=1}^{C} \sum_{i=1}^{H} \sum_{j=1}^{W} ( T_{k,i,j}^{l} \\
&- \mathcal{G}\left(f_{align}(S_{k,i,j}^{l}) \cdot M_{i,j}^{l} \right) )^2,
\end{split}
\end{equation}
where \(L\) is the number of layers for distillation, \(C\), \(H\), and \(W\) denote the channel, height, and width of the feature map, respectively, \(S\) and \(T\) represent the feature maps of the student and teacher models, \(f_{align}(\cdot)\) is a feature transformation function that aligns the student features to match the dimensions of the teacher features, \(M\) is a randomly generated spatial binary mask used to select a subset of features for distillation, and \(\mathcal{G}(\cdot)\) is a stop-gradient operation that prevents gradients from flowing through the masked features~\cite{yang2022maskedgenerativedistillation}.

\subsubsection{Kullback-Leibler Divergence Loss}
KL divergence loss is a classical and widely used method for knowledge distillation. Unlike CWD and MGD, which focus more on the intermediate feature layer, KL loss focuses on the predictive distributions of the output layers of the teacher model and the student model. The idea is to use KL divergence loss to measure the difference between the two outputs after converting them into probability distributions by means of a softmax function~\cite{hinton2015distillingknowledgeneuralnetwork}.
This approach encourages the student model to learn not only the final predictions of the teacher model, but also to understand the relative confidence of the teacher model for each category. KL divergence loss is simple to implement and computationally efficient, and is often used as the basis for knowledge distillation. However, since it only focuses on the distribution of the final output layer, it makes relatively little use of the deeper knowledge of the internal features of the teacher model, which may not be captured as comprehensively as CWD and MGD.
\begin{equation}
\mathcal{L}_{\text{KL}} = \tau^2 \cdot \text{KL}\left( \text{Softmax}\left( \frac{z_T}{\tau} \right) \parallel \text{Softmax}\left( \frac{z_S}{\tau} \right) \right),
\label{equ:kl}
\end{equation}
where \( z_T \) and \( z_S \) are the teacher and student logits, \( \tau \) is the temperature, and \( \mathrm{KL}(\cdot \| \cdot) \) denotes the KL divergence.

\section{Experiments and discussion}

We used the COCO2017 dataset~\cite{lin2015microsoftcococommonobjects}, and the Phase-to-Space (P2S) Transform~\cite{mao2021acceleratingatmosphericturbulencesimulation} is used to generate synthetic turbulence, simulating atmospheric distortions such as blurring, aberrations, and intensity fluctuations. Specifically, each image from COCO2017 is transformed by P2S to produce a video containing 50 frames. In total, the training set comprises 4,594 samples, and the validation set contains 503 samples. There are total of 69 object categories.

\subsection{AT mitigation model selection}

Three representative turbulence mitigation methods are selected for comparison in this task: TMT~\cite{zhang2023imagingatmosphereusingturbulence}, DATUM~\cite{zhang2024spatiotemporalturbulencemitigationtranslational} and MAMAT~\cite{hill2025mamat3dmambabasedatmospheric}. During training, input sequences were randomly cropped to 256$\times$256 patches to enhance spatial diversity. We trained the models for 200 epochs using the Adam optimizer with a fixed learning rate of 0.0001. A sliding window strategy was employed, where five neighboring frames were used as input to predict the central frame, applied consistently during both training and inference. Following \cite{anan2023atmospheric}, this configuration provides an optimal balance and achieves the best PSNR performance. Two common image quality metrics are used: Peak Signal-to-Noise Ratio (PSNR) and Structural Similarity (SSIM).

Table~\ref{tab:restoration_quality} shows the average image quality metrics, including PSNR and SSIM, of the different turbulence mitigation methods. From the results, it can be seen that all the restoration methods have achieved a significant improvement in both metrics when compared to the original turbulent image (Distorted), which is a good enough indication of the effectiveness of these restoration methods in mitigating atmospheric turbulence.

\begin{table}[t]
\centering
\caption{Performance of different AT  mitigation methods.}
\begin{tabular}{l|cc}
\hline
\textbf{Method} & \textbf{PSNR}$\uparrow$ & \textbf{SSIM}$\uparrow$ \\
\hline
Distorted & 20.85 & 0.51 \\
DATUM~\cite{zhang2024spatiotemporalturbulencemitigationtranslational} & 22.77 & 0.64 \\
TMT~\cite{zhang2023imagingatmosphereusingturbulence} & 22.96 & 0.65 \\
MAMAT~\cite{hill2025mamat3dmambabasedatmospheric} & \textbf{23.97} & \textbf{0.66} \\
\hline
\end{tabular}
\label{tab:restoration_quality}
\end{table}

\begin{table}[t]
\centering
\small
\caption{Detection performance and model size of different architectures.}
\begin{tabular}{l|cc}
\hline
\textbf{Method} & \textbf{mAP@50:95}$\uparrow$ & \textbf{\# Param (M)}$\downarrow$ \\
\hline
RetinaNet~\cite{Lin_2017_ICCV} & 0.22 & 34.0 \\
DETR~\cite{carion2020endtoendobjectdetectiontransformers} & 0.23 & 41.3 \\
LW-DETR medium~\cite{chen2024lwdetrtransformerreplacementyolo} & 0.36 & 28.2 \\
YOLOv11 large~\cite{khanam2024yolov11overviewkeyarchitectural} & \textbf{0.44} & \textbf{25.3} \\
\hline
\end{tabular}
\label{tab:detection_comparison}

\end{table}

\begin{table*}
\centering
\caption{Performance comparison on distillation models}
\begin{tabular}{l|ccccc}
\hline
\textbf{Model} & \textbf{param
(M)}$\downarrow$ & \textbf{PSNR}$\uparrow$ & \textbf{SSIM}$\uparrow$ & \textbf{mAP (\%)}$\uparrow$ & \textbf{Time (ms)}$\downarrow$ \\
\hline
MAMAT+YOLOv11-L (Teacher) & 2.8+25.3 & 23.40 & 0.66 & 44.11 & 70.9 \\
Separate Distillation (Students) & 0.6+2.6 & 23.37 & 0.61 & 33.18 & 37.6 \\
Joint Distillation (JDATT-Small) & 0.6+2.6 & 23.43 & 0.63 & 33.50 & 35.7 \\
Joint Distillation (JDATT-Medium) & 0.7+9.4 & 23.56 & 0.62 & 34.56 & 40.7 \\
Joint Distillation (JDATT-Large) & 2.5+20.1 & 23.63 & 0.64 & 39.26 & 54.6 \\
\hline
\end{tabular}
\label{tab:model_comparison}
\end{table*}

\subsection{Target detection model selection}
\label{sec:Detetion_Selection}

Achieving real-time object detection in atmospheric turbulence conditions requires a careful trade-off between detection accuracy, model size, and inference speed. Using the restored data from MAMBA, we evaluate a selection of mainstream detection frameworks:  
RetinaNet~\cite{Lin_2017_ICCV},  DETR~\cite{carion2020endtoendobjectdetectiontransformers}, LW-DETR~\cite{chen2024lwdetrtransformerreplacementyolo} and YOLOv11~\cite{khanam2024yolov11overviewkeyarchitectural}. Each framework offers multiple model variants of different complexities; for fairness, we select one variant from each with comparable parameter counts to enable meaningful comparisons.

Object detection performance is evaluated using the standard COCO-style Average Precision (AP) metric, calculated as the area under the precision-recall curve across multiple Intersection over Union (IoU) thresholds ranging from 0.5 to 0.95 in 0.05 increments. The mean AP across all thresholds, denoted as mAP@50:95, is reported as the primary metric. Table~\ref{tab:detection_comparison} shows that YOLOv11-Large performs best on the restored data by MAMAT in terms of both precision and parameter count.

\begin{figure}
    \begin{center}
    \includegraphics[trim={0 0  17.5cm 0}, clip,width=0.9\columnwidth]{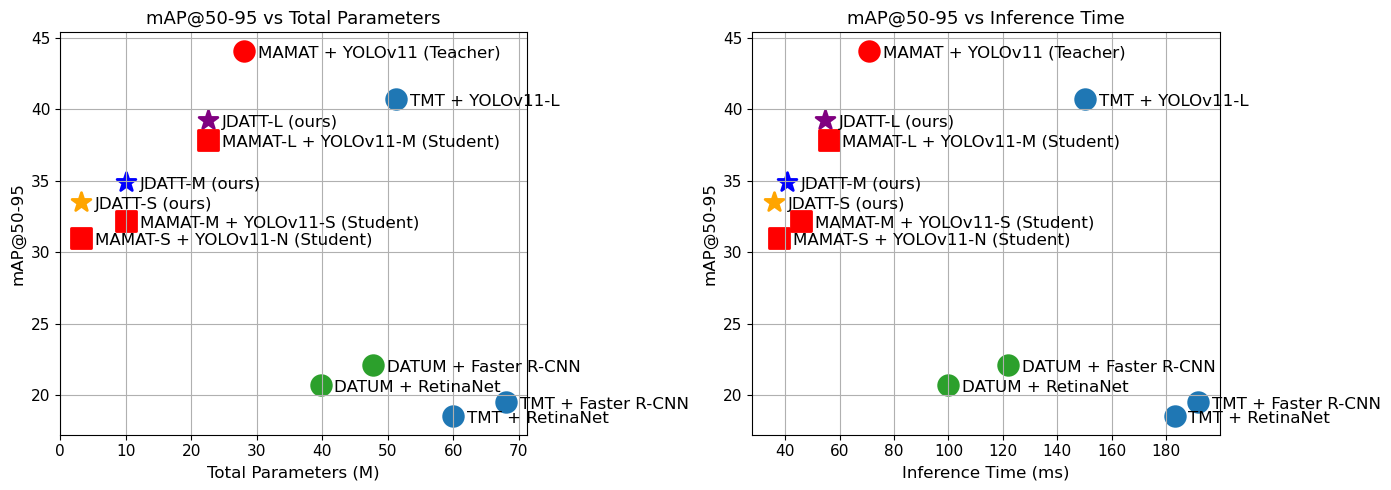}
    \includegraphics[trim={17.5cm 0  0 0}, clip,width=0.9\columnwidth]{images/performanceplot.png}
    \end{center} 
    \caption{Performance comparison of object detection under atmospheric turbulence condition. Our models are marked with stars, where S, M, and L denote small, medium, and large models, respectively. Square markers represent student models trained separately, while circle markers indicate original models used for both the atmospheric turbulence mitigator and the detector. }
    \label{fig:output1}
\end{figure}
\begin{figure*}
    \begin{center}
    \includegraphics[width=\linewidth]{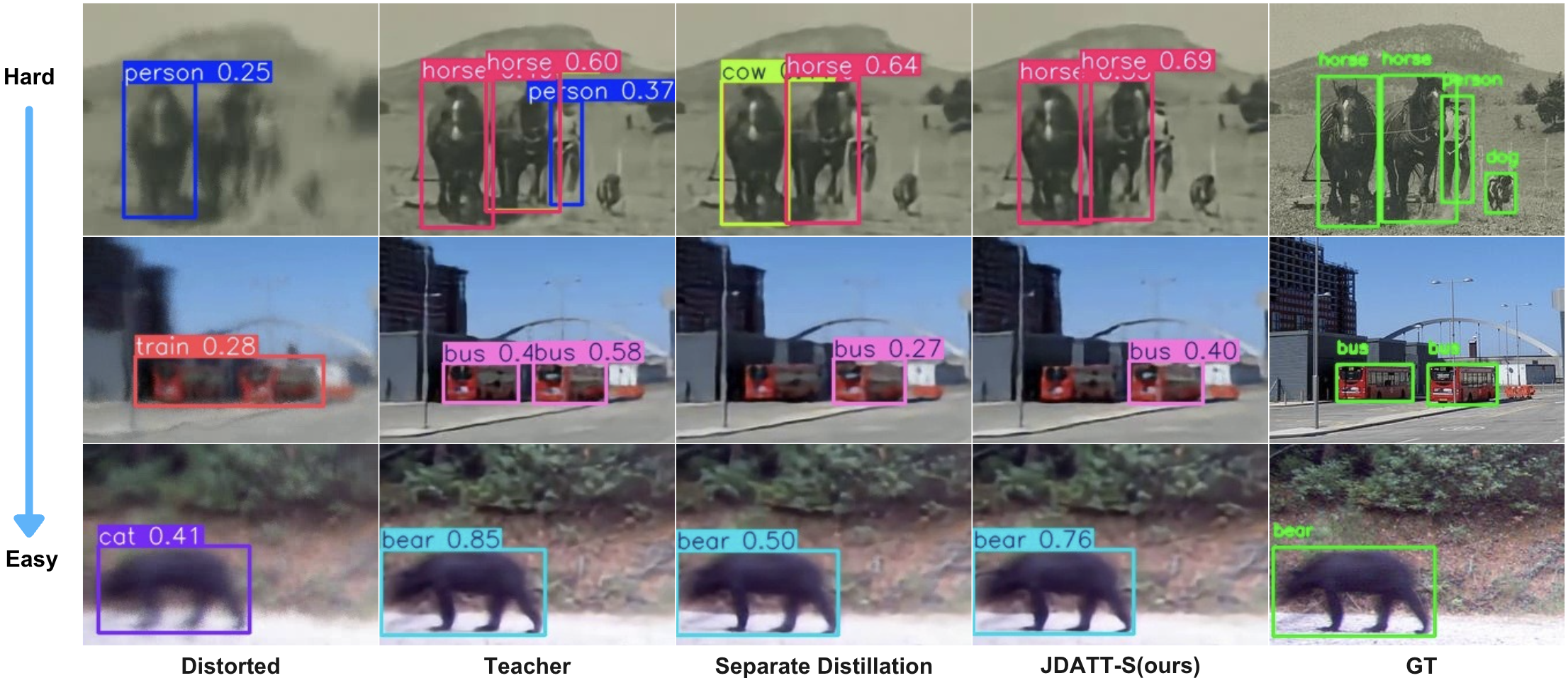}
    \end{center} 
    \caption{Results on synthetic atmospheric turbulence sequences of varying difficulty levels. In hard cases, some models exhibit missed or incorrect detections, whereas in easier scenarios, all models successfully detect targets, though with varying confidence scores.}
    \label{fig:synresults}
\end{figure*}
\begin{figure*}
    \begin{center}
    \includegraphics[width=\linewidth]{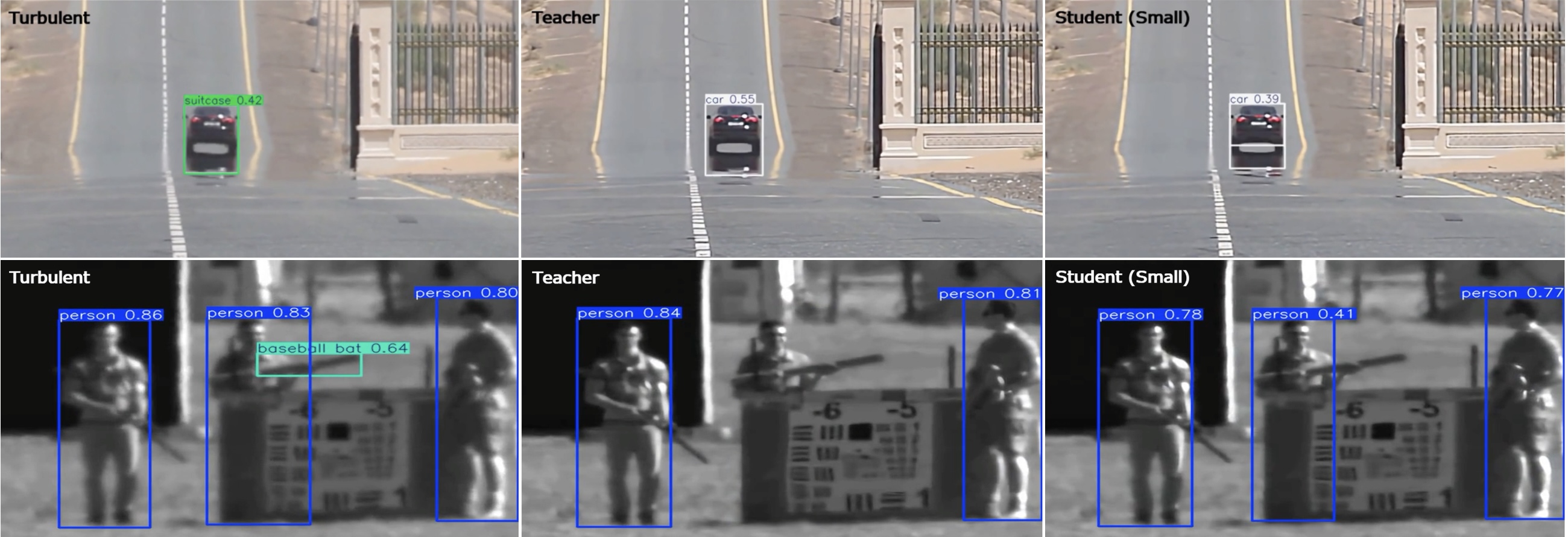}
    \end{center} 
    \caption{Results on real atmospheric turbulence data, comparing the original distorted videos with outputs from both teacher and student models.}
    \label{fig:results_realAT}
\end{figure*}

\subsection{Joint knowledge distillation}

We adopt MAMAT and YOLOv11-Large as fixed teacher models for atmospheric turbulence mitigation and target detection, respectively. For student models, we design three variants of increasing complexity: small (S), medium (M), and large (L). Each student AT mitigator is a lightweight adaptation of the original MAMAT architecture, preserving its two-stage design—registration (to compensate for pixel shifts) and enhancement (to restore sharpness and contrast)—while significantly reducing computational overhead. Key modifications include the removal of deformable convolutions, a reduction in the number of 3D convolutional layers and feature maps (halved in the latter case), and the retention of fewer Res-Mamba blocks along with a single Transformer module in the refinement stage.  These changes maintain core restoration capabilities while improving efficiency. The S and M student architectures are visualized in Fig.~\ref{fig:distillation} (right). The L model increases depth in both the registration and enhancement stages to improve capacity and retains the original number of feature map channels from MAMAT. Overall, model sizes are reduced from 2.8M (original MAMAT) to 2.5M (L), 0.7M (M), and 0.6M (S), respectively.

For the student detection models, we adopt architecture variants provided by Ultralytics based on YOLOv11, specifically YOLOv11-M, YOLOv11-S, and YOLOv11-N. Compared to the original YOLOv11-L model (25.3M parameters), the student models are significantly more compact, with model sizes of 20.1M (L), 9.4M (M), and 2.6M (S), respectively. During training, input patches of size 256$\times$256 are used with a batch size of 1 and gradient accumulation over 4 steps. The learning rate is initialized at 1.5e-4 and follows a cosine annealing schedule with 10 warm-up epochs. We apply a weight decay of 0.01, and the training runs for 100 epochs in total.

Table~\ref{tab:model_comparison} and Fig.~\ref{fig:output1} present both the visual restoration quality and detection accuracy achieved by different methods and model variants. Notably, joint distillation leads to improved atmospheric turbulence mitigation performance, even as model size is significantly reduced. In contrast, target detection exhibits a more noticeable drop in precision with smaller model variants. Nonetheless, models trained via joint distillation consistently outperform those trained with separate stage-wise distillation. For the smallest model, PSNR, SSIM, and mAP improve by 0.26\%, 3.28\%, and 0.96\%, respectively. Compared to the teacher model, the student achieves an 88.6\% reduction in model size, with a slight improvement in PSNR and a 24.1\% drop in mAP.

Fig.~\ref{fig:synresults} and Fig.~\ref{fig:results_realAT} show qualitative results on both synthetic and real atmospheric turbulence data, respectively. For the synthetic dataset, ground truth labels are available and provided for comparison. Both teacher and student models demonstrate improved visual clarity by effectively mitigating turbulence distortions. Notably, models trained via joint distillation exhibit more accurate target detection compared to those trained with separate distillation. This is evident in challenging synthetic video cases and real AT sequences, where separately distilled models either misdetect or miss objects, while the jointly trained models perform more robustly.


\section{Conclusion}

This paper presents a novel knowledge distillation framework, JDATT, designed to reduce model size and improve inference speed for visual enhancement and object detection under atmospheric turbulence conditions. We propose a joint end-to-end training strategy that preserves image quality using a combination of reconstruction loss, Channel-Wise Distillation (CWD) loss, and Masked Generative Distillation (MGD) loss. Detection performance is maintained through detection loss and Kullback–Leibler (KL) divergence. Experimental results demonstrate that our framework achieves substantial model compression while maintaining performance comparable to the teacher models. Moreover, the proposed joint training strategy consistently outperforms conventional separate distillation approaches.

\bibliographystyle{unsrt}
\bibliography{egbib}
\end{document}